\documentclass[runningheads]{llncs}

\usepackage{eccv}

\usepackage{eccvabbrv}
\usepackage{enumitem}
\usepackage{graphicx}
\usepackage{booktabs}
\usepackage{Yin}
\usepackage{bm} 
\usepackage{pifont}
\usepackage{multirow}
\usepackage{arydshln}
\usepackage{array}
\usepackage{colortbl}
\usepackage{xcolor}
\usepackage{wrapfig}
\usepackage{algorithmic}
\usepackage{tcolorbox}
\usepackage{algorithm}
\usepackage[accsupp]{axessibility}  
\usepackage{makecell}
\usepackage{hhline}
\newcommand{\pub}[1]{{\color{gray}{\tiny{[{#1}]}}}}
\newcommand{\pubbig}[1]{{\color{gray}{\tiny{[{#1}]}}}}

\usepackage{marvosym}

\usepackage{hyperref}
\definecolor{cvprblue}{rgb}{0.21,0.49,0.74}
\definecolor{tabname}{rgb}{0.92, 0.92, 0.92}
\usepackage{orcidlink}

\begin{document}

\title{\textsc{Cut-ViT}: Task-Specific Model Pruning via Gram Anchoring Subspace Consistency} 

\titlerunning{Cut-ViT}

\author{Jianjian Yin\inst{1,4,*} \orcidlink{0000-0002-0445-454X} \and
Liulei Li\inst{2,*} \orcidlink{0000-0002-4637-0328} \and
Tao Chen\inst{1,4} \orcidlink{0000-0001-8239-1698} \and Yi Chen\inst{3} \orcidlink{0000-0002-8762-4523} \and \\ Yazhou Yao\inst{1,4}\textsuperscript{(\Letter)} \orcidlink{0000-0002-0337-9410} \and Wenguan Wang\inst{2}  \orcidlink{0000-0002-0802-9567}}

\authorrunning{J.~Yin et al.}

\institute{ $^1$Nanjing University of Science and Technology, Nanjing, China \\ ~  $^2$Zhejiang University, Hangzhou, China.  
$^3$Nanjing Normal University, Nanjing, China  \\
$^4$State Key Laboratory of Intelligent Manufacturing of Advanced
Construction Machinery, Nanjing, China \\ 
$^{*}$Equal contribution \quad
$\textsuperscript{\Letter}$Corresponding author \\
\href{https://github.com/NUST-Machine-Intelligence-Laboratory/Cut-ViT}{\textcolor[rgb]{0.80,0.00,0.25}{https://github.com/NUST-Machine-Intelligence-Laboratory/Cut-ViT}} 
}

\maketitle

\begin{abstract}
  Pruning visual foundation models has attracted considerable attention. However, existing methods focus on rigid point-to-point token alignment on a single dataset for pruning, suffering from  two limitations:  \textbf{i}) robustness degradation, and \textbf{ii}) task-specificity deficiency.
  To address these limitations, we propose a task-specific pruning pipeline, named \textsc{Cut-ViT}.
  Specifically, we first construct gram anchoring matrices from both spatial and semantic perspectives, and perform the subspace decomposition to extract the corresponding subspace bases. Basis-agnostic and residual constraints are then adopted to align the gram subspaces between the native and pruned DINOv3 models along spatial and channel dimensions, enabling subnetworks to inherit robust feature representations of native DINOv3.  Furthermore, we design spectral entropy adaptation, which quantifies the information density of feature manifolds along spatial and channel dimensions, thereby adapting the pruning objective to specific downstream tasks.
  Experiments show that \textsc{Cut-ViT} requires approximately \textbf{one} minute on a single A100 GPU to obtain subnetworks at various sparsity levels, using only \textbf{20.9\%} of the time and \textbf{45.5\%} of the GPU memory compared with previous methods, while achieving SOTA performance on \textbf{six} tasks across \textbf{nine} datasets. 
  
\end{abstract}

\section{Introduction}\label{sec:introduction}

The emergence of visual foundation models (VFMs)~\cite{kirillov2023segment,caron2021emerging,oquab2024dinov2,simeoni2025dinov3}, notably the DINO series~\cite{caron2021emerging,oquab2024dinov2,simeoni2025dinov3}, has pushed the boundaries of visual representation learning \cite{zhou2026learning,ma2026learning,Yin_2026_CVPR,li2023semantic,zhou2025unialign,sun2025jo, cai2026iris}. However, despite VFMs achieving competitive performance across a wide range of downstream tasks~\cite{zhou2024prototype,YinCPLYNH25,gu2026medfg,yang2026beyond,cai2026unbiased,li2026reconciling}, their massive computational overhead poses significant challenges for deployment on resource-constrained edge devices. Consequently, the efficient compression of VFMs without compromising performance has become a research focus~\cite{zhou2025attention,li2025frequency,haberer2024hydravit,zhu2025ea,zhang2024slicing}. Among various model compression paradigms, training-free one-shot structured pruning (OSP)~\cite{simoncini2025elastic,LucasM25,kohama2023single,lee2021layer} has recently emerged as a highly promising solution. By dynamically ranking parameter importance using loss gradients,
OSP enables rapid generation of lightweight subnetworks, making it particularly suitable for large-scale VFMs where retraining is computationally prohibitive.

\begin{figure}[t]
	\centering
	\includegraphics[width=1\linewidth]{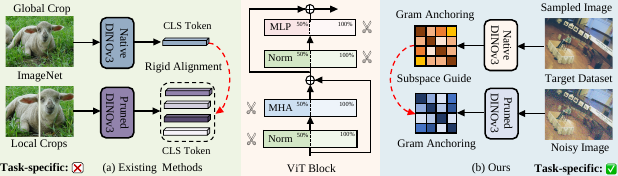}
	\caption{Comparison with prior work. (a) To ensure feature consistency, existing methods enforce rigid alignment between CLS tokens; (b) \textsc{Cut-ViT} aligns the subspace induced by the inherent gram anchoring within DINOv3, thereby enabling the subnetworks to inherit  the robust feature representations of the original network.}
	\label{fig1_motivation}
	\vspace{-0.3cm}
\end{figure}

However, existing OSP methods face two bottlenecks that hinder their applications in real-world scenarios. 
\ding{182} \textbf{Robustness degradation}. Current OSP methods rely on rigid point-to-point alignment between the tokens of native and pruned models. This localized strategy encourages the subnetwork to memorize specific numerical features rather than preserving the global semantic topology. 
By prioritizing exact token matching over structural consistency, these methods fail to inherit the robust manifold structure of VFMs, leading to representation degradation. \ding{183} \textbf{Task-specificity deficiency}. Prior training-free methods typically follow a rigid pipeline, that optimizes on a single generic dataset (\eg, ImageNet) to deploy a universal subnetwork across all downstream tasks (Fig. \ref{fig1_motivation}(a)). 
While seemingly elegant, this paradigm overlooks that  for edge computing, model capacity  has already been sacrificed  to achieve faster inference and lower memory usage.
Emphasizing a shared, task-agnostic  subnetwork inevitably leads to a further compromise in precision and results in a sub-optimal trade-off between precision and efficiency for specific downstream applications.

To address these challenges, this work proposes \textsc{Cut-ViT}, a task-specific OSP algorithm that can generate subnetworks at various sparsity levels in approximately \textbf{\textit{one minute}} for DINOv3~\cite{simeoni2025dinov3} using a single A100 GPU. 
To alleviate \ding{182}, we  rethink feature alignment from the perspective of manifold consistency. Previous studies~\cite{simeoni2025dinov3} found that the robustness of DINOv3 stems from the rich global topology encoded in its gram anchoring, which acts as a dense regularization constraint.
However, rigid point-to-point alignment fails to preserve this structure and is sensitive to noise. 
Inspired by this, a gram anchoring subspace decomposition strategy (Fig. \ref{fig1_motivation}(b)) is proposed. 
Rather than matching raw features, \textsc{Cut-ViT} constructs spatial and channel gram matrices and decomposes them to extract orthonormal bases. These bases span the directions of maximal joint variation, thereby decoupling the core semantic topology from high-frequency noise, which enables the design of a basis-invariant subspace consistency strategy. By imposing basis-agnostic and residual constraints, \textsc{Cut-ViT}  forces the  feature subspace of subnetworks to geometrically align with that of the teacher. This allows the subnetwork to inherit the robust spatial and semantic manifold of DINOv3 without overfitting to specific token values.
To answer \ding{183}, we design a pruning pipeline that adapts to both data distributions and task properties.  \textbf{First}, unlike generic pruning, we perform optimization directly on the target dataset. By encoding the target data distribution into the gram anchoring and propagating it via subspace consistency, the subnetwork implicitly adapts to domain shifts without requiring ground-truth annotations.
\textbf{Second},  \textsc{Cut-ViT} introduces spectral entropy to quantify the information density of feature manifolds.
Since different tasks require different information granularities (\eg, classification relies on global semantics, while segmentation demands high spatial frequency)~\cite{woo2023convnext,he2022masked}, spectral entropy measures this complexity along spatial and channel dimensions. This serves to dynamically weight the pruning objective to tailor the model architecture for specific downstream tasks.

\textsc{Cut-ViT} demonstrates superior performance over existing training-free OSP methods~\cite{kohama2023single,LucasM25,simoncini2025elastic} across \textbf{nine} datasets covering \textbf{six}  downstream tasks: semantic segmentation, object detection, video object segmentation, semantic matching, depth estimation, and image classification.
Beyond standard benchmarks, our out-of-distribution experiments further confirm that \textsc{Cut-ViT}  can preserve the \text{robustness} of the native DINOv3~\cite{simeoni2025dinov3} model. 
In terms of \textit{efficiency}, \textsc{Cut-ViT} offers a compelling advantage.  Specifically, it achieves a \textbf{3.4\%} improvement in $(\mathcal{J}\&\mathcal{F})_m$ for video object segmentation on DAVIS-2017~\cite{pont20172017}, \textbf{5.1\%} improvement in PCK@0.05 for semantic matching on FG3DCar~\cite{lin2014jointly}, \textbf{2.8\%} improvement in mAP for object detection on MS COCO~\cite{lin2014microsoft} compared with prior state-of-the-art OSP approaches, while requiring only \textbf{20.9\%} of the inference time and \textbf{45.5\%} of the GPU memory. Moreover, our approach remains competitive even against training-based model pruning alternatives~\cite{haberer2024hydravit,zhu2025ea}, while consuming only \textbf{0.44\%} of the pruning time and \textbf{28.2\%} of the GPU memory.

\section{Related Work}

\textbf{Visual Foundation Model.} The emergence of large-scale pretraining~\cite{xiao2025visual,xiao2025prompt,yang2024depth} and the success of visual encoders~\cite{DosovitskiyB0WZ21,liu2021swin,touvron2021training} have catalyzed the development of VFMs, such as CLIP~\cite{radford2021learning}, SAM~\cite{kirillov2023segment}, and DINO~\cite{caron2021emerging,oquab2024dinov2,simeoni2025dinov3}. Trained on vast datasets, these models learn robust and generalized visual representations that are transferable across a wide array of vision tasks \cite{qu2026self,yang2026efficiency,xu2026gsv2x,yang2025changetitans,cai2026unbiased,qu2026robust,yin2026depmatch,li2024human}. However, their considerable computational overhead creates significant challenges for deployment on  edge devices. Several studies have aimed to lightweight VFMs, including Efficientsam~\cite{xiong2024efficientsam}, Theia~\cite{shang25a}, and Tinysam~\cite{shu2025tinysam}. These methods primarily rely on knowledge distillation to transfer the capabilities of computationally intensive encoders to more efficient, lightweight encoders. In contrast, our approach selects parameters directly from the VFM’s encoder, generating networks with varying sparsities in a single round of inference.

\noindent\textbf{Model Compression.}  The limitation of computational resources has sparked increased interest in model compression research~\cite{zhou2025attention,kim2025random,li2025frequency,lee2025customkd}. In recent years, quantization~\cite{gholami2025casp,rastegari2016xnor,he2023ptqd}, token optimization~\cite{kim2025faster,chen2024image,zeng2025token,lee2024multi,bolya2022token,luo2025tr}, and structured pruning~\cite{kwon2022fast,haberer2024hydravit} have become the primary research directions in this field. Quantization~\cite{wang2025quest,wang2024q} aims to reduce the precision of model weights, thus striving to minimize computational overhead while  preserving model performance. Token optimization~\cite{chen2024efficient,bergner2025token} pruns redundant tokens and merges tokens with high similarity, leading to acceleration in model computation. Structured pruning~\cite{Sun0BK24,sanh2020movement}, which eliminates network parameters, has inspired a series of influential works. HydraViT~\cite{haberer2024hydravit} generates multiple sub-networks by stacking attention during training, progressively modifying the embedding dimension and the  head number of attention layer. Scala~\cite{zhang2024slicing} employs scale-aware coordination during training to ensure that each pruned network learns robust objectives. EA-ViT~\cite{zhu2025ea} adopts a two-stage training paradigm and incorporates a lightweight router to select sub-networks that meet budget constraints. However, these methods require intensive training, leading to high computational cost.

\vspace{0.05cm}

\noindent\textbf{One-shot Pruning without Retraining.} One-shot structured pruning~\cite{frantar2023sparsegpt,kwon2022fast,yang2023global} eliminates the need for training, requiring only a single inference pass, during which parameter importance is dynamically ranked based on gradients derived from the loss function. This allows for the generation of networks with varying sparsities in a single pass. Specifically, LAMP~\cite{lee2021layer} introduces a layer-adaptive pruning score that enables pruning without requiring hyperparameter tuning, while SNIP~\cite{kohama2023single} employs a magnitude-based criterion. SNOWS \cite{LucasM25} and SnapViT \cite{simoncini2025elastic} are notable OSP methods designed for post-processing pretrained models. The former introduces a second-order optimization framework combining Hessian-free optimization with a customized conjugate gradient method for layer-wise pruning, effectively addressing second-order approximation issues. The latter jointly evaluates parameter importance for pruning using a self-supervised objective and a genetic algorithm.  The aforementioned methods focus on rigid point-to-point alignment between tokens for pruning, which undermines the robustness of the subnetworks. Conversely, our work encourages alignment of gram anchoring in subspace, ensuring that the subnetworks inherit the robust feature representations of the native DINOv3.

Departing from prior SVD-based work \cite{wang2025svd,chen2025accelerating} that rely on \textit{static} subspace construction, Cut-ViT pioneers rotation-invariant \textit{dynamic}  optimization to align the subspaces of the teacher and pruned models during pruning, ensuring the inheritance of robust topological structures.

\section{Methodology}
\label{method}

\subsection{Preliminary}
\label{pre}

Standard training-free OSP methods~\cite{lee2021layer,kohama2023single,frantar2023sparsegpt} aim to identify an optimal binary mask $\bm{m} \in \{0,1\}^{|\bm{W}|}$ for the weights $\bm{W}$ of a pre-trained VFM (\textit{e.g.,} DINOv3~\cite{simeoni2025dinov3}) to minimize the performance gap between the pruned student and the native teacher. 
Typically, this is formulated as a reconstruction problem:
\begin{equation}
	\label{eq:osp}
	H \approx \frac{1}{N} \sum\nolimits_{n=1}^{N} 
	\left\| \nabla_{\bm{m}} \mathcal{L}_n \right\|^2 ,
	\qquad 
    \mathcal{L} = \Upsilon(\bm{O}_s, \bm{O}_t).
    \qquad
\end{equation}
The parameter importance score $H$ is approximated using the empirical Fisher matrix over $N$ calibration samples. 
Based on these scores, search algorithms such as xNES\cite{glasmachers2010exponential} are employed to locate the optimal subnetwork. $\bm{O}_s$ and $\bm{O}_t$ denote the feature outputs of the student (local crops) and teacher (global crops), respectively. 
$\Upsilon(\cdot,\cdot)$ represents the distillation objective, traditionally implemented as a rigid element-wise comparison (\eg, MSE or KL divergence). 


\begin{figure}[t]
	\centering
	\includegraphics[width=1\linewidth]{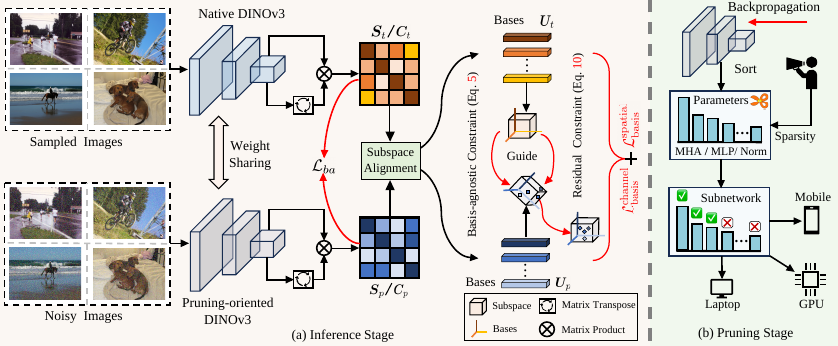}
	\caption{The pipeline of our \textsc{Cut-ViT}. Pairs of images are input into the native and pruning-guided DINOv3  to obtain feature embeddings. $\Lcal_{ba}$ denotes the dense feature matching loss. Subspace decomposition ($\S\ref{GSC}$) is applied to the spatial and channel gram matrices ($\bm{S}_t$, $\bm{S}_p$, $\bm{C}_t$, $\bm{C}_p$) to derive subspace bases ($\bm{U}_t$, $\bm{U}_p$). Subspace consistency  is enforced with basis-agnostic  and residual  constraints ($\S\ref{SAR}$), yielding spatial and channel gram losses, $\mathcal{L}^\text{spatial}_\text{basis}$ and $\mathcal{L}^\text{channel}_\text{basis}$. During pruning, backpropagation computes the gradients of each parameter, which are then ranked in descending order of magnitude to select parameters and generate subnetworks with varying sparsity levels.}
	\label{fig2_arch}
\end{figure}


\subsection{Gram Anchoring Subspace Decomposition}
\label{GSC}
\textbf{Motivation.} 
The strong generalization ability of DINOv3\!~\cite{simeoni2025dinov3} arises from the rich structural and semantic relationships encoded in its dense feature representations. While prior work \cite{simoncini2025elastic,LucasM25} enforce rigid point-wise alignment of tokens, such exact numerical reconstruction imposes overly strict constraints that can encourage overfitting to redundant correlations and high-frequency noise. 
Therefore, 
we propose  decomposing the gram matrix into subspaces spanned by dense features using singular value decomposition (SVD), and then regularizing the subspace consistency for gram anchoring.  The detailed pipeline of \textsc{Cut-ViT} is shown in Fig. \ref{fig2_arch}. By injecting noise into input images, the derived bases are encouraged to capture a more robust and noise-invariant semantic topology.

.

\noindent \textbf{Gram Formulation.} Let $\bm{F}\in\mathbb{R}^{L\times D}$  denote a feature embedding, where $L$ is the  token number and $D$ is the channel dimension. We define a generalized gram matrix $\bm{G}$ to model the second-order correlation of the features along a specific dimension, and then extract the manifold structure by performing SVD on $\bm{G}$:

\begin{equation}
	\label{eq:svd_general}
	\bm{G} = \bm{X} \cdot \bm{X}^\top \approx \bm{U} \bm{\Sigma} \bm{U}^\top,
\end{equation}
where $\bm{X}$ represents the feature matrix unfolded along the target dimension, $\bm{\Sigma}$
is the diagonal matrix of singular values, and $\bm{U}$
contains the orthonormal basis vectors spanning the dominant subspace. This formulation allows us to capture correlations from two complementary perspectives:

\noindent $\bullet$ \ \textbf{Spatial Gram Anchoring.} 
To capture the topological relationships between tokens (\eg, the spatial layout of objects), we compute the spatial gram matrix $\bm{S}$
by aggregating features across the channel dimension, yielding:
\begin{equation}
	\label{eq:spatial} 
	\bm{S} = \bm{F} \cdot \bm{F}^\top \in \mathbb{R}^{L \times L}.
\end{equation} 
After applying Eq.\!~\ref{eq:svd_general} to $\bm{S}$, the spatial basis matrix $\bm{U}^{\bm{S}}$ can be obtained,  which identifies the dominant modes of inter-token correlation and can group spatially coherent patches based on their feature similarity (``Where'').

\noindent $\bullet$ \  \textbf{Channel Gram Anchoring.} 
Complementary to spatial structure, channel correlations encode semantic attributes (\eg, texture, style, and class). We compute the channel gram matrix $\bm{C}$
by treating tokens as samples, resulting in:
\begin{equation}
	\label{eq:channel}
	\bm{C} = \bm{F}^\top \cdot \bm{F} \in \mathbb{R}^{D \times D}.
\end{equation}
Similarly, decomposing $\bm{C}$
yields the channel basis $\bm{U}^{\bm{C}}$, 
which decouples abstract concepts from spatial instantiations. Consequently, $\bm{U}^{\bm{C}}$ serves as a compact descriptor of the global semantics, and answers the question of ``What'' is depicted.  Detailed process is provided  

\begin{wrapfigure}[8]{r}{0.62\textwidth}  
	\centering
	\vspace{-1.28cm}
	\includegraphics[width=0.62\textwidth]{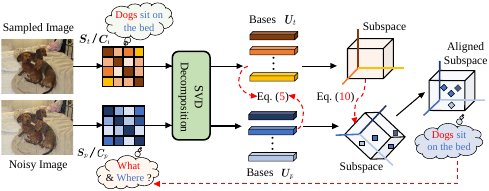}  
	\vspace{-20pt}
	\caption{The process of spatial and channel gram anchoring subspace consistency.}
	\label{fig:gram}
\end{wrapfigure}
\noindent in Fig. \ref{fig:gram}. By extracting $\bm{U}^{\bm{S}}$ and $\bm{U}^{\bm{C}}$, the alignment between teacher and student models is converted into a subspace consistency problem, so as to preserve   the robust structural and semantic topology of DINOv3 \!~\cite{simeoni2025dinov3}.

\subsection{Basis-Invariant Subspace Consistency}
\label{SAR}
This section addresses the challenge of aligning feature subspaces of pruned and native models. Since the singular vectors (bases) derived by SVD are not unique, directly aligning the basis vectors element-by-element via MSE loss is ill-posed. To tackle this, we propose a basis-agnostic constraint that aligns the subspaces independent of the specific basis orientation. A residual constraint is further proposed to suppress noise outside the aligned subspace.

\noindent\textbf{Basis-agnostic  Constraint.} 
Let $\bm{U}_p\in\mathbb{R}^{L\times K}$ and $\bm{U}_t\in\mathbb{R}^{L\times K}$ denote the truncated Top-$K$ basis matrices for the pruned and native models. The affinity matrix is defined as $\bm{M}=\bm{U}_p^\top\bm{U}_t$. The subspace consistency is enforced by maximizing the subspace overlap (\ie, correlation energy) between two sets of bases:
\vspace{-0.15cm}
\begin{equation}
	\label{eq5}
	\Lcal_\text{basis} = 1 - ||\bm{M}||_{F}^{2} = 1 - \sum\nolimits_{i=1}^{K}\sum\nolimits_{j=1}^{K} {M}_{ij}^{2},
	\vspace{-0.15cm}
\end{equation}
where $||\cdot||_{F}^{2}$ denotes the Frobenius norm, and $K$ is the number of selected components. 
$\Lcal_\text{basis}$ is robust to the non-uniqueness of SVD, as established below.

\vspace{0.2cm}
\begin{tcolorbox}[
	colback=gray!10,
	boxrule=0pt,
	colframe=gray!10,
	arc=2pt, 
	boxsep=0pt,
	]
		\textbf{Theorem. \textit{(Basis Invariance)}.} The loss function $\Lcal_\text{basis}$
		depends solely on the subspace spanned by the bases and is invariant to arbitrary orthogonal transformations of the basis vectors $\bm{U}_p$
		and $\bm{U}_t$.
\end{tcolorbox} 

\begin{proof}
	Let $\bm{Q}_p, \bm{Q}_t \in \mathbb{R}^{K\times K}$ be arbitrary orthogonal matrices (\ie, $\bm{Q}_p^\top\bm{Q}_p=\bm{I}$). The non-uniqueness of SVD implies that valid bases can be transformed versions $\tilde{\bm{U}}_p=\bm{U}_p\bm{Q}_p$ and $\tilde{\bm{U}}_t=\bm{U}_t\bm{Q}_t$. We examine the invariance of $\Lcal_\text{basis}$ in two cases:
	
	\noindent \textbf{Case 1: Simultaneous Transformation.}
	Consider the scenario where both the native and pruned bases undergo arbitrary rotations. Substituting the transformed bases into the Frobenius norm term yields:
	\begin{equation}
		||\tilde{\bm{U}}_p^\top \tilde{\bm{U}}_t||_F^2 = ||(\bm{U}_p \bm{Q}_p)^\top (\bm{U}_t \bm{Q}_t)||_F^2 = ||\bm{Q}_p^\top (\bm{U}_p^\top \bm{U}_t) \bm{Q}_t||_F^2. 
	\end{equation}
	Recalling the unitary invariance property of the Frobenius norm
	(\ie, $||\bm{U}\bm{A}\bm{V}||_{{F}}=||\bm{A}||_{F}$ for orthogonal 
	$\bm{U}, \bm{V}$), we have:
	\begin{equation} 
		||\bm{Q}_p^\top \bm{M} \bm{Q}_t||_F^2 = ||\bm{M}||_F^2.
	\end{equation}
	
	\noindent \textbf{Case 2: Unilateral Transformation.}
	In our task-specific pruning setting, the native DINOv3 serves as a frozen teacher, whose basis $\bm{U}_t$
	is fixed ($\bm{Q}_t=\bm{I}$). The basis $\bm{U}_p$ of the pruned network evolves during optimization and may rotate arbitrarily. Therefore,  the objective becomes:
	\begin{equation}
		||\tilde{\bm{U}}_p^\top \bm{U}_t||_F^2 = ||\bm{Q}_p^\top (\bm{U}_p^\top \bm{U}_t)||_F^2 = ||\bm{Q}_p^\top \bm{M}||_F^2.
	\end{equation}
	Since $\bm{Q}_p^\top$ is orthogonal, the multiplication represents an isometric rotation which preserves the Frobenius norm, \ie, $||\bm{Q}_p^\top\bm{M}||_F=||\bm{M}||_F$. Details are provided in \textit{supplementary materials}.
	Thus, $\Lcal_\text{basis}$
	consistently aligns the subspace of the pruned model to the teacher, 
	regardless of how pruned bases are oriented. 
	\qed
\end{proof}

\noindent \textbf{Residual Constraint.} While $\Lcal_\text{basis}$ ensures the alignment of principal directions, the pruned feature embedding 
$\bm{F}_p$ inevitably contains redundant information orthogonal to the target subspace. To explicitly filter this noise, we employ a geometric projection. Let $\bm{P}_t=\bm{U}_t\bm{U}_t^\top$ be the orthogonal projection matrix onto the native DINOv3 subspace. Any feature embedding 
$\bm{F}_p$ from the pruned network can be uniquely decomposed into two orthogonal components:
\begin{equation}
	\label{eq8}
	\bm{F}_p =
	\underbrace{\left(\bm{U}_t \bm{U}_t^\top \bm{F}_p\right)}_{\text{Aligned Component}}
	+ ~~~~
	\underbrace{\left((\bm{I} - \bm{U}_t \bm{U}_t^\top)\bm{F}_p\right)}_{\text{Orthogonal Residual}}.
\end{equation}
The first term represents features successfully projected onto the robust teacher subspace. The second term corresponds to the residual error or noise lying outside the target manifold. We minimize the energy of irrelevant information:
\begin{equation}
	\label{eq9}
	\mathcal{L}_\text{residual} = ||(\bm{I} - \bm{U}_t \bm{U}_t^\top) \bm{F}_p||_{F}^{2}.
\end{equation}
$\mathcal{L}_\text{residual}$ suppresses activation that does not structurally align with the gram anchoring of the teacher model, thereby enhancing feature purity.

\subsection{Task-specific Pruning via Spectral Entropy Adaptation}
\label{TAP} 

In this section, we inject tailored properties for each task into the training-free pruning pipeline.
\textbf{First}, rather than performing universal pruning on a generic dataset (\textit{e.g.,} ImageNet \cite{deng2009imagenet}), which ignores the distinct data distributions, our pruning is conducted on a minimal subset of $N$ samples  from the target dataset. 
This  allows the subnetwork to capture the underlying data manifold of the target domain without requiring annotations. It is noteworthy that, benefited from our efficient design (\ie, 61s \textit{v.s.} 292s of SnapViT \cite{simoncini2025elastic}), \textsc{Cut-ViT} maintains lower  latency than prior methods even when performing  multiple searches. 
\textbf{Second}, we introduce spectral entropy to adaptively weight the pruning objective based on the information density of feature manifolds
across spatial and channel axes.
Let $\sigma_i^{\bm{S}}$ and $\sigma_j^{\bm{C}}$ denote the singular values of the native spatial and channel gram matrices.
We first normalize these singular values to obtain the energy probability distributions within the spatial and channel subspaces:
\begin{equation}
p_i^{\bm{S}} = \frac{\sigma_i^{\bm{S}}}{\sum_k \sigma_k^{\bm{S}}}, \ \ \ p_j^{\bm{C}} = \frac{\sigma_j^{\bm{C}}}{\sum_k \sigma_k^{\bm{C}}}.
\end{equation}
The complexity of spatial and channel representations can then be measured as:
\begin{equation}
\textstyle \mathbb{H}(\bm{S}_t) = - \sum_i p_i^{\bm{S}} \log p_i^{\bm{S}},\ \ \ \mathbb{H}(\bm{C}_t) = - \sum_j p_j^{\bm{C}} \log p_j^{\bm{C}}
\end{equation}
Theoretically, the eigenspectrum decay of the feature covariance matrices reflects the degree of representation collapse within the token space.
For instance, classification tasks optimize for shift-invariant global semantics, which drives spatial tokens toward uniformity. This concentrates the spatial Gram matrix energy into a few \textbf{dominant principal components} (lower $\mathbb{H}(\bm{S}_t)$). 
Conversely, dense prediction tasks (\textit{e.g.,} semantic segmentation, correspondence matching) should preserve high-frequency topological boundaries and pixel-level semantics. 
This motivates a penalty in representation collapse, and enforces a much \textbf{flatter singular value distribution} along the spatial axis (higher $\mathbb{H}(\bm{S}_t)$).
On this basis, we remodel the basis consistency loss to be task-aware:
\begin{equation}
\mathcal{L}^\text{all}_\text{basis} = w^\text{spatial} \cdot \mathcal{L}^\text{spatial}_\text{basis} + w^\text{channel} \cdot \mathcal{L}^\text{channel}_\text{basis},
\end{equation}
where the dynamic weights are defined as:
\begin{equation}
w^\text{spatial} = \frac{\mathbb{H}(\bm{S}_t)}{\mathbb{H}(\bm{S}_t) + \mathbb{H}(\bm{C}_t)}, \ \ \ w^\text{channel} = \frac{\mathbb{H}(\bm{C}_t)}{\mathbb{H}(\bm{S}_t) + \mathbb{H}(\bm{C}_t)}.
\end{equation}
This ensures that our pruning algorithm adapts to given downstream tasks \textit{w.r.t.} the underlying representational demands.

\section{Experiments}
\label{experiment}
This section presents a comprehensive evaluation of \textsc{Cut-ViT} across multiple visual tasks at sparsity levels ranging from 10\% to 30\%.  
The datasets used for each task are listed below.
\vspace{0.05cm}

\noindent\textbf{Datasets.} We use the following datasets for various tasks: ADE20K \cite{zhou2017scene} and PASCAL VOC 2012 \cite{everingham2015pascal} for semantic segmentation, COCO \cite{lin2014microsoft} for object detection, NYUv2 \cite{silberman2012indoor} for depth estimation, DAVIS-2017 \cite{pont20172017} for video object segmentation, and FG3DCAR \cite{lin2014jointly}, JODS \cite{rubinstein2013unsupervised}, and SBD \cite{hariharan2011semantic} for semantic matching, as well as ImageNet \cite{deng2009imagenet} for image classification.
ADE20K is a large semantic segmentation dataset containing 150 categories, with 20,210/2,000/3,000 images allocated for train/val/test. PASCAL VOC 2012  consists of 20 semantic classes and one background class, with 1,464/1,449/1,456 images used for train/val/test.  COCO has 80 object categories for object detection, with 118,287 training images, 5,000 validation images, and 40,000 test images.  NYUv2 is split into 795 images for training and 654 images for testing. DAVIS-2017 contains 60 training videos and 30 validation videos. The three datasets for semantic matching together comprise 400 pairs of test images with dense correspondence annotations. ImageNet consists of 1,000 categories, with 1.2 million images allocated for training, 50,000 for validation, and 100,000 for testing.

\noindent\textbf{Model Architecture.} We adopt DINOv3 \cite{simeoni2025dinov3} (ViT-B/16) as the encoder for all tasks. Detailed parameter counts and FLOPs at different sparsity levels are provided in the supplementary material. Object detection uses a standard deformable attention decoder \cite{Xizhou_Zhu}. For classification, semantic segmentation, and depth estimation, we employ linear probing. Video object segmentation and semantic matching are evaluated via feature matching, without any decoder.
\vspace{0.05cm} 

\noindent\textbf{Evaluation Metrics.} The performance of classification, semantic segmentation, object detection, and semantic matching is evaluated using Top-1 Accuracy, mIoU, mAP with IoU thresholds in the range [0.5 : 0.05 : 0.95], and PCK@0.05, respectively. Depth estimation is assessed using ARel (lower is better) and $\delta_1$ (higher is better), while video object segmentation performance is measured by mean region similarity $\Jcal_m$, mean contour-based accuracy $\Fcal_m$, and $(\Jcal \& \Fcal)_m$.
\vspace{0.05cm}

\noindent\textbf{Implementation Details.} The model is implemented using the PyTorch framework. Pruning is performed on a single  GPU, while all other experiments are conducted on four NVIDIA A100 GPUs. 
For semantic segmentation, the learning rate is set to 1e-3, with training conducted over 100 epochs. For object detection, the learning rate is  5e-4, and the model is trained for 72 epochs. For depth estimation, the learning rate is 1e-6, with training spanning 60 epochs. The batch size is set to 16 for all tasks. For image classification, the learning rate is 5e-3, the batch size is 1024, and the model is trained for 20 epochs. Video object segmentation and semantic matching tasks do not require training; models pruned at various sparsity levels are directly validated on the datasets.
We set $N = 1000$ and the number of principal components $K = 192$ in the low-rank SVD. Except for image classification, which relies solely on the CLS token for basis-invariant subspace consistency, all other downstream tasks incorporate basis-agnostic and residual constraints from the perspectives of spatial and channel gram anchoring.


\begin{table*}[t]
	\begin{center}
		\renewcommand\arraystretch{1.15}
		\caption{Quantitative results ($\S\ref{compare}$) on DAVIS-2017 \cite{pont20172017} video object segmentation using {\scriptsize $(\Jcal\&\Fcal)_m$ / $\Jcal_m$ / $\Fcal_m$}.}
		\vspace{-0.4cm}
		\setlength{\tabcolsep}{0.2pt}
		\resizebox{1.0\textwidth}{!}{
			{\scriptsize
				\begin{tabular}{>{\raggedleft}p{3.4cm}>{\centering\arraybackslash}p{1.6cm}||c|c|c}
					\thickhline
					\rowcolor[rgb]{0.92,0.92,0.92} &  &      \multicolumn{3}{c}{Different Sparsity Levels of DINOv3} \\ \hhline{>{\arrayrulecolor{tabname}}--|>{\arrayrulecolor{black}}---}
					\rowcolor[rgb]{0.92,0.92,0.92} \multicolumn{1}{c}{\multirow{-2}{*}{Method}}  & \multicolumn{1}{c||}{\multirow{-2}{*}{Dataset}}   & 30\% & 20\% & 10\% \\ 
					\hline \hline
					\multicolumn{5}{l}{\textit{\textbf{Training-based  (0\% : 69.7 / 67.00 / 72.5)}}} \\[0.5ex]
					\hline 
					HydraViT \pubbig{NeurIPS24} \cite{haberer2024hydravit} & DAVIS & 58.6 / 58.1 / 59.1 & 64.5 / 63.1 / 65.9 & 68.4 / 66.1 / 70.7 \\
					EA-ViT  \pubbig{ICCV25} \cite{zhu2025ea}   & DAVIS & 60.3 / 59.4 / 61.2 & 65.4 / 63.9 / 66.9 & 69.3 / 66.8 / 71.8  \\ \hline \hline
					
					\multicolumn{5}{l}{\textit{\textbf{Training-free (0\% : 69.7 / 67.0 / 72.5)}}} \\[0.5ex]
					\hline 
					LAMP  \pubbig{ICLR21} \cite{lee2021layer}  &ImageNet  & 35.0 / 32.9 / 37.0 & 40.4 / 38.1 / 42.8 & 60.2 / 58.4 / 62.1 \\
					NVIT \pubbig{CVPR23} \cite{yang2023global} & ImageNet & 36.3 / 34.2 / 38.4 & 48.3 / 46.9 / 49.7 & 61.2 / 59.7 / 62.7  \\
					SNIP \pubbig{ICCV23} \cite{kohama2023single}  & ImageNet  & 36.5 / 33.8 / 39.2 & 51.0 / 49.2 / 52.9 & 61.9 / 60.3 / 64.4  \\
					SNOWS \pubbig{ICLR25} \cite{LucasM25}   & ImageNet & 52.9 / 50.7 / 55.1  & 59.4 / 57.8 / 61.0 & 63.8 / 61.3 / 66.3  \\
					SnapViT \pubbig{NeurIPS25} \cite{simoncini2025elastic}  & ImageNet  & 56.0 / 54.3 / 57.7 & 60.8 / 58.8 / 62.8 & 65.0 / 63.4 / 66.7 \\
					\cdashline{1-5}
					\noalign{\vskip 1pt}
					
					\rowcolor[rgb]{0.92,0.92,0.92}	 \textbf{\textsc{Cut-ViT} (Ours)}~~~    & DAVIS  & \makecell{\textbf{59.4} / \textbf{58.0} / \textbf{60.8} \\ {\tiny \textbf{\textcolor{impro}{$\uparrow$3.4}}}~/ {\tiny \textbf{\textcolor{impro}{$\uparrow$3.7}}} / {\tiny \textbf{\textcolor{impro}{$\uparrow$3.1}}}}  & \makecell{\textbf{65.0} / \textbf{63.2} / \textbf{66.8} \\ {\tiny \textbf{\textcolor{impro}{$\uparrow$4.2}}}~/ {\tiny \textbf{\textcolor{impro}{$\uparrow$4.4}}} / {\tiny \textbf{\textcolor{impro}{$\uparrow$4.0}}}}  & \makecell{\textbf{68.7}  / \textbf{66.3} / \textbf{71.2} \\ {\tiny \textbf{\textcolor{impro}{$\uparrow$3.7}}}~/ {\tiny \textbf{\textcolor{impro}{$\uparrow$2.9}}} / {\tiny \textbf{\textcolor{impro}{$\uparrow$4.5}}}} \\
					
					\hline
				\end{tabular}
			}
		}
		\label{tab:video_object_segmentation}
		\vspace{-0.4cm}
	\end{center}
\end{table*}

\begin{table*}[t]
	\begin{center}
		\renewcommand\arraystretch{1.15}
		\caption{Quantitative results ($\S\ref{compare}$) of semantic matching on the FG3DCar / JODS / SBD (FJS) dataset \cite{lin2014jointly,rubinstein2013unsupervised,hariharan2011semantic}.}
		\vspace{-0.4cm}
		\setlength{\tabcolsep}{1.0pt}
		\resizebox{1.0\textwidth}{!}{
			{\scriptsize
				\begin{tabular}{>{\raggedleft}p{3.4cm}>{\centering\arraybackslash}p{1.6cm}||c|c|c}
					\thickhline
					\rowcolor[rgb]{0.92,0.92,0.92} &  &     \multicolumn{3}{c}{Different Sparsity Levels of DINOv3} \\  \hhline{>{\arrayrulecolor{tabname}}--|>{\arrayrulecolor{black}}---}
					\rowcolor[rgb]{0.92,0.92,0.92} \multicolumn{1}{c}{\multirow{-2}{*}{Method}}  & \multicolumn{1}{c||}{\multirow{-2}{*}{Dataset}}   & 30\% & 20\% & 10\% \\ 
					\hline \hline
					\multicolumn{5}{l}{\textit{\textbf{Training-based  (0\% : 92.0 / 77.1 / 59.7)}}} \\[0.5ex]
					\hline 
					HydraViT \pubbig{NeurIPS24} \cite{haberer2024hydravit} & FJS & 69.9 / 55.7 / 37.6 & 90.7 / 72.5 / 51.9 & 91.2 / 76.3 / 59.1 \\
					EA-ViT \pubbig{ICCV25} \cite{zhu2025ea} & FJS & 72.0 / 57.9 / 38.5 & 91.4 / 74.2 / 54.6 & 92.0 / 77.0 / 59.5\\ \hline \hline
					\multicolumn{5}{l}{\textit{\textbf{Training-free (0\% : 92.0 / 77.1 / 59.7)}}} \\[0.5ex]
					\hline 
					LAMP \pubbig{ICLR21} \cite{lee2021layer}  &ImageNet   & 55.2 / 38.8 / 29.9 & 79.5 / 56.9 / 38.9 & 84.2 / 59.2 / 50.4 \\
					NVIT \pubbig{CVPR23} \cite{yang2023global} & ImageNet & 56.3 / 38.2 / 29.8  & 80.2 / 56.4 / 39.2 & 86.3 / 60.7 / 51.1  \\
					SNIP \pubbig{ICCV23} \cite{kohama2023single}  & ImageNet   & 59.6 / 39.1 / 30.4 & 82.3 / 58.9 / 40.9 & 88.4 / 63.6 / 53.2 \\
					SNOWS \pubbig{ICLR25} \cite{LucasM25}   & ImageNet & 64.8 / 44.2 / 32.1  & 85.3 / 62.9 / 50.6 & 90.3 / 71.7 / 54.9  \\
					SnapViT \pubbig{NeurIPS25} \cite{simoncini2025elastic} & ImageNet  & 65.7 / 43.3 / 32.4   & 86.0 / 64.0 / 50.8  & 90.2 / 75.1 / 56.0   \\
					\cdashline{1-5}
					\noalign{\vskip 1pt}
					\rowcolor[rgb]{0.92,0.92,0.92}  \textbf{\textsc{Cut-ViT} (Ours)}~~~    & FJS  & \makecell{\textbf{70.8} / \textbf{56.3} / \textbf{37.9} \\ {\tiny \textbf{\textcolor{impro}{$\uparrow$5.1}}}~/ {\tiny \textbf{\textcolor{impro}{$\uparrow$12.0}}} / {\tiny \textbf{\textcolor{impro}{$\uparrow$5.5}}}} & \makecell{\textbf{90.6} / \textbf{73.0} / \textbf{53.2} \\ {\tiny \textbf{\textcolor{impro}{$\uparrow$4.6}}}~/ {\tiny \textbf{\textcolor{impro}{$\uparrow$9.0}}} / {\tiny \textbf{\textcolor{impro}{$\uparrow$2.4}}}}  & \makecell{\textbf{92.0} / \textbf{76.9} / \textbf{58.9} \\ {\tiny \textbf{\textcolor{impro}{$\uparrow$1.7}}}~/ {\tiny \textbf{\textcolor{impro}{$\uparrow$1.8}}} / {\tiny \textbf{\textcolor{impro}{$\uparrow$2.9}}}} \\
					
					\hline
				\end{tabular}
			}
		}
		\label{tab:semantic_matching}
		\vspace{-0.4cm}
	\end{center}
\end{table*}

\subsection{Comparison with SOTA Methods}
\label{compare}

\noindent\textbf{Video Object Segmentation.} We compare \textsc{Cut-ViT} with other state-of-the-art methods on the DAVIS-2017 dataset \cite{pont20172017} for video object segmentation, as shown in Table \ref{tab:video_object_segmentation}. The proposed method significantly outperforms the training-free method on the $(\mathcal{J} \& \mathcal{F})_m$, $\mathcal{J}_m$, and $\mathcal{F}_m$ metrics at various pruning sparsity levels. Notably, at 20\% sparsity, \textsc{Cut-ViT} improves the corresponding metrics by over \textbf{4.2\%}, \textbf{4.4\%}, and \textbf{4.0\%}, respectively. When compared to training-based pruning methods, our approach also outperforms HydraViT \cite{haberer2024hydravit}, although it slightly trails behind EA-ViT \cite{zhu2025ea} in performance. These results highlight the superiority of the backbone obtained using our pruning approach.

\noindent\textbf{Semantic Matching.} Table \ref{tab:semantic_matching} reports the semantic matching results on the FJS dataset \cite{lin2014jointly,rubinstein2013unsupervised,hariharan2011semantic}. Across all datasets and pruning sparsity levels, \textsc{Cut-ViT} consistently outperforms existing training-free methods as well as the training-based HydraViT \cite{haberer2024hydravit}. For instance, at 30\% sparsity, \textsc{Cut-ViT} improves performance over the training-free methods by \textbf{5.1\%}, \textbf{12.1\%}, and \textbf{5.5\%} on FG3DCar, JODS, and PASCAL, respectively, and exceeds HydraViT \cite{haberer2024hydravit} by 0.9\%, 0.6\%, and 0.3\%. These results indicate that the proposed pruning strategy effectively preserves the fine-grained  feature matching capability of DINOv3 \cite{simeoni2025dinov3}.

\begin{table*}[t]
	\centering
	\renewcommand\arraystretch{1.15}
	\setlength{\tabcolsep}{0.1pt}
	
	\begin{minipage}[t]{0.488\textwidth}
		\centering
		\caption{Quantitative results ($\S\ref{compare}$) for object detection on COCO \cite{lin2014microsoft}.}
		\vspace{-0.2cm}
		\resizebox{\textwidth}{!}{
			\begin{tabular}{r>{\centering\arraybackslash}p{1.7cm}||>{\centering\arraybackslash}p{1.37cm}|
					>{\centering\arraybackslash}p{1.37cm}|
					>{\centering\arraybackslash}p{1.37cm}}
				\spthickhline
				\rowcolor[rgb]{0.92,0.92,0.92} &  &      \multicolumn{3}{c}{\normalsize  Different Sparsities} \\  \hhline{>{\arrayrulecolor{tabname}}--|>{\arrayrulecolor{black}}---}
				\rowcolor[rgb]{0.92,0.92,0.92} \multicolumn{1}{c}{\multirow{-2}{*}{\normalsize Method}}  & \multicolumn{1}{c||}{\multirow{-2}{*}{\normalsize Dataset}}   & {\normalsize     30\%} & {\normalsize 20\%} & {\normalsize 10\%} \\ 
				\hline \hline
				\multicolumn{5}{l}{\textit{\textbf{\normalsize Training-based  (0\% : 57.8)}}} \\[0.5ex]
				\hline  
				{\normalsize HydraViT}\pub{NeurIPS24} \cite{haberer2024hydravit}  & {\normalsize COCO} & {\normalsize 47.6} & {\normalsize 52.3} & {\normalsize 56.1}\\ 
				{\normalsize EA-ViT}\pub{ICCV25} \cite{zhu2025ea}  & {\normalsize COCO} & {\normalsize 49.4} & {\normalsize 54.1} & {\normalsize 56.9}\\ \hline \hline
				
				\multicolumn{5}{l}{\textit{\textbf{\normalsize Training-free  (0\% : 57.8)}}} \\[0.5ex]
				\hline
				{\normalsize LAMP}\pub{ICLR21} \cite{lee2021layer}  & {\normalsize ImageNet}  &  {\normalsize 42.3} & {\normalsize 43.8} & {\normalsize 50.6} \\
				{\normalsize NVIT}\pubbig{CVPR23}  \cite{yang2023global} & {\normalsize ImageNet}  & {\normalsize 42.7} & {\normalsize 44.6} & {\normalsize 51.4}  \\
				{\normalsize SNIP}\pub{ICCV23} \cite{kohama2023single}  & {\normalsize ImageNet}  & {\normalsize 43.8} & {\normalsize 45.2} & {\normalsize 48.4}  \\
				{\normalsize SNOWS}\pub{ICLR25} \cite{LucasM25}   & {\normalsize ImageNet} & {\normalsize 45.1}  & {\normalsize 50.6} & {\normalsize 54.4} \\
				{\normalsize SnapViT}\pub{NeurIPS25} \cite{simoncini2025elastic}  & {\normalsize ImageNet}  & {\normalsize 45.8} & {\normalsize 51.0} &  {\normalsize 54.1} \\ 
				\cdashline{1-5}
				\noalign{\vskip 1pt}
				\rowcolor[rgb]{0.92,0.92,0.92} {\normalsize \textbf{\textsc{Cut-ViT} (Ours)}}   & {\normalsize COCO}  & \textbf{\normalsize 48.6} {\scriptsize\textbf{\textcolor{impro}{$\uparrow$2.8}}} & \textbf{\normalsize 53.0} {\scriptsize\textbf{\textcolor{impro}{$\uparrow$2.0}}}  & \textbf{\normalsize 56.5} {\scriptsize \textbf{\textcolor{impro}{$\uparrow$2.1}}} \\
				\hline
			\end{tabular}
			\label{tab:object_detection}
		}
	\end{minipage}
	\hfill
	\begin{minipage}[t]{0.488\textwidth}
		\centering
		\caption{Quantitative results ($\S\ref{compare}$) for semantic segmentation on ADE20K \cite{zhou2017scene}.}
		\vspace{-0.2cm}
		\resizebox{\textwidth}{!}{
			\begin{tabular}{r>{\centering\arraybackslash}p{1.7cm}||>{\centering\arraybackslash}p{1.35cm}|
					>{\centering\arraybackslash}p{1.35cm}|
					>{\centering\arraybackslash}p{1.35cm}}
				\spthickhline
				\rowcolor[rgb]{0.92,0.92,0.92} &    &     \multicolumn{3}{c}{\normalsize Different Sparsities} \\  \hhline{>{\arrayrulecolor{tabname}}--|>{\arrayrulecolor{black}}---}
				\rowcolor[rgb]{0.92,0.92,0.92} \multicolumn{1}{c}{\multirow{-2}{*}{\normalsize Method}}  & \multicolumn{1}{c||}{\multirow{-2}{*}{\normalsize Dataset}}   & {\normalsize 30\%} & {\normalsize 20\%} & {\normalsize 10\%} \\ 
				\hline \hline
				\multicolumn{5}{l}{\textit{\textbf{\normalsize Training-based  (0\% : 51.8)}}} \\[0.5ex]
				\hline 
				{\normalsize HydraViT} \pub{NeurIPS24} \cite{haberer2024hydravit}  & {\normalsize ADE20K} & {\normalsize 46.4} & {\normalsize 48.9} & {\normalsize 49.6}\\ 
				{\normalsize EA-ViT} \pub{ICCV25} \cite{zhu2025ea}  & {\normalsize ADE20K} & {\normalsize 48.2} & {\normalsize 50.9} & {\normalsize 51.4} \\ \hline \hline
				\multicolumn{5}{l}{\textit{\textbf{\normalsize Training-free  (0\% : 51.8)}}} \\[0.5ex]
				\hline
				{\normalsize LAMP} \pub{ICLR21} \cite{lee2021layer}   & {\normalsize ImageNet}  &  {\normalsize 43.7} & {\normalsize 45.4} & {\normalsize 47.3} \\
				{\normalsize NVIT} \pubbig{CVPR23}  \cite{yang2023global} & {\normalsize ImageNet}  & {\normalsize 43.5} & {\normalsize 46.1} & {\normalsize 47.9}  \\
				{\normalsize SNIP} \pub{ICCV23} \cite{kohama2023single}  & {\normalsize ImageNet}  & {\normalsize 44.0} & {\normalsize 47.6} & {\normalsize 48.0}  \\
				{\normalsize SNOWS} \pub{ICLR25} \cite{LucasM25}   & {\normalsize ImageNet} & {\normalsize 45.8}  & {\normalsize 47.8} & {\normalsize 50.0} \\
				{\normalsize SnapViT} \pub{NeurIPS25} \cite{simoncini2025elastic}  & {\normalsize ImageNet}  & {\normalsize 46.0} & {\normalsize 48.3} &  {\normalsize 50.2} \\ 
				\cdashline{1-5}
				\noalign{\vskip 1pt}
				\rowcolor[rgb]{0.92,0.92,0.92} {\normalsize \textbf{\textsc{Cut-ViT} (Ours)}}   & {\normalsize ADE20K}  & \textbf{\normalsize 47.2}{\scriptsize\textbf{\textcolor{impro}{$\uparrow$1.2}}} & \textbf{\normalsize 50.0}{\scriptsize\textbf{\textcolor{impro}{$\uparrow$1.7}}}  & \textbf{\normalsize 51.0}{\scriptsize \textbf{\textcolor{impro}{$\uparrow$0.8}}} \\
				\hline
			\end{tabular}
			\label{tab:semantic_segmentation}
		}
	\end{minipage}
\end{table*}

\noindent\textbf{Object Detection.} Table \ref{tab:object_detection} compares the performance of \textsc{Cut-ViT} with state-of-the-art methods on the COCO dataset \cite{lin2014microsoft}. \textsc{Cut-ViT} outperforms training-free approaches and achieves competitive performance relative to the training-based EA-ViT \cite{zhu2025ea}. For instance, at 30\% sparsity, \textsc{Cut-ViT} improves mAP by \textbf{2.8\%} over the former, while slightly trailing behind the latter by 0.4\%. These results further demonstrate that our approach more effectively preserves the detection capabilities of DINOv3 \cite{simeoni2025dinov3}.

\noindent\textbf{Semantic Segmentation.} Table \ref{tab:semantic_segmentation} presents experimental results comparing our method with SOTA approaches on the ADE20K dataset \cite{zhou2017scene}. At sparsity levels of 30\%, 20\%, and 10\%, \textsc{Cut-ViT} outperforms training-free methods by \textbf{1.2\%}, \textbf{1.7\%}, and \textbf{0.8\%} in mIoU, respectively, and significantly surpasses the training-based HydraViT \cite{haberer2024hydravit}. These results show that \textsc{Cut-ViT} preserves robust semantic information in subnetworks pruned from DINOv3 \cite{simeoni2025dinov3}.

\noindent\textbf{Depth Estimation.}  Table \ref{tab:depth_estimation} presents the depth estimation results on the NYUv2 dataset \cite{silberman2012indoor}, comparing the proposed method with state-of-the-art approaches. Consistent with its performance on other dense prediction tasks, \textsc{Cut-ViT} outperforms training-free pruning methods in both ARel and $\delta_1$, while achieving competitive performance compared to training-based methods. Furthermore, when pruning an equal number of parameters, our approach yields models whose performance is closer to the original DINOv3 \cite{simeoni2025dinov3} than those produced by training-free pruning methods. These results across multiple dense prediction tasks demonstrate that \textsc{Cut-ViT} effectively preserves the robust dense feature representations of the native DINOv3 \cite{simeoni2025dinov3} during pruning.

\begin{table*}[t]
	\centering
	\renewcommand\arraystretch{1.15}
	\setlength{\tabcolsep}{1.0pt}
	\begin{minipage}[t]{0.55\textwidth}
		\centering
		\renewcommand\arraystretch{1.15}
		\caption{Quantitative results ($\S\ref{compare}$) for  depth estimation on  NYUv2 \cite{silberman2012indoor}  using ARel / $\delta_1$.}
		\vspace{-0.2cm}
		\resizebox{\textwidth}{!}{
			\begin{tabular}{r>{\centering\arraybackslash}p{1.8cm}||>{\centering\arraybackslash}p{1.65cm}|
					>{\centering\arraybackslash}p{1.65cm}|
					>{\centering\arraybackslash}p{1.65cm}}
				\spthickhline
				\rowcolor[rgb]{0.92,0.92,0.92} &    &    \multicolumn{3}{c}{\normalsize Different Sparsities} \\  \hhline{>{\arrayrulecolor{tabname}}--|>{\arrayrulecolor{black}}---}
				\rowcolor[rgb]{0.92,0.92,0.92} \multicolumn{1}{c}{\multirow{-2}{*}{\normalsize Method}}  & \multicolumn{1}{c||}{\multirow{-2}{*}{\normalsize Dataset}}   & {\normalsize 30\%} & {\normalsize 20\%} & {\normalsize 10\%} \\ 
				\hline \hline
				\multicolumn{5}{l}{\textit{\textbf{\normalsize Training-based  (0\% : 3.0 / 99.9)}}} \\[0.5ex]
				\hline 
				{\normalsize HydraViT} \pub{NeurIPS24} \cite{haberer2024hydravit}  & {\normalsize NYUv2} & {\normalsize 5.2 / 98.4} & {\normalsize 4.9 / 99.0} & {\normalsize 4.1 / 99.7}\\ 
				{\normalsize EA-ViT} \pub{ICCV25} \cite{zhu2025ea}  & {\normalsize NYUv2} & 3.9 / 99.3 & {\normalsize 3.4} / {\normalsize 99.7} & {\normalsize 3.1} / {\normalsize 99.9}\\ \hline \hline
				\multicolumn{5}{l}{\textit{\textbf{\normalsize Training-free  (0\% : 3.0 / 99.9)}}} \\[0.5ex]
				\hline
				{\normalsize LAMP} \pub{ICLR21} \cite{lee2021layer} & {\normalsize ImageNet}  & {\normalsize 15.8} / {\normalsize 91.8} & {\normalsize 14.0} / {\normalsize 93.4} & {\normalsize 11.3} / {\normalsize 96.4} \\
				{\normalsize NVIT} \pubbig{CVPR23}  \cite{yang2023global} & {\normalsize ImageNet}  & {\normalsize 14.4 / 93.3} & {\normalsize 12.7 / 95.1} & {\normalsize 10.3 / 96.1}  \\
				{\normalsize SNIP} \pub{ICCV23} \cite{kohama2023single}  & {\normalsize ImageNet}  & {\normalsize 11.1} / {\normalsize 95.0} & {\normalsize 10.4} / {\normalsize 96.3} & {\normalsize 9.2} / {\normalsize 96.8}  \\
				{\normalsize SNOWS} \pub{ICLR25} \cite{LucasM25}   & {\normalsize ImageNet} & {\normalsize 8.3} / {\normalsize 97.6}  & {\normalsize 5.7} / {\normalsize 98.2} & {\normalsize 5.5} / {\normalsize 98.1}  \\
				{\normalsize SnapViT} \pub{NeurIPS25} \cite{simoncini2025elastic}  & {\normalsize ImageNet}  & {\normalsize 7.7} / {\normalsize 97.6} & {\normalsize 5.6} / {\normalsize 98.5} &  {\normalsize 5.0} / {\normalsize 99.2} \\ 
				\cdashline{1-5}
				\noalign{\vskip 1pt}
				\rowcolor[rgb]{0.92,0.92,0.92}{\normalsize \textbf{\textsc{Cut-ViT} (Ours)}}   & {\normalsize NYUv2}  & \makecell{\textbf{4.6} / \textbf{98.7} \\ {\scriptsize\textbf{\textcolor{impro}{$\uparrow$3.1}}} / {\scriptsize\textbf{\textcolor{impro}{$\uparrow$1.1}}}}  & \makecell{\textbf{3.8} / \textbf{99.3} \\ {\scriptsize\textbf{\textcolor{impro}{$\uparrow$1.8}}} / {\scriptsize\textbf{\textcolor{impro}{$\uparrow$0.8}}}}  & \makecell{\textbf{\normalsize 3.3}  / \textbf{\normalsize 99.9} \\ {\scriptsize\textbf{\textcolor{impro}{$\uparrow$1.7}}} / {\scriptsize\textbf{\textcolor{impro}{$\uparrow$0.7}}}}   \\
				\hline
			\end{tabular}
			\label{tab:depth_estimation}
		}
	\end{minipage}
	\hfill
	\begin{minipage}[t]{0.43\textwidth}
		\renewcommand\arraystretch{1.185}
		\centering
		\caption{Quantitative  results ($\S\ref{ood}$) for out of distribution  validation.}
		\vspace{-0.2cm}
		\resizebox{\textwidth}{!}{
			\begin{tabular}{r>{\centering\arraybackslash}p{1.7cm}||>{\centering\arraybackslash}p{1.35cm}|
					>{\centering\arraybackslash}p{1.35cm}}
				\spthickhline
				\rowcolor[rgb]{0.92,0.92,0.92} &    &    \multicolumn{2}{c}{\normalsize ADE20K $\rightarrow$ VOC} \\  \hhline{>{\arrayrulecolor{tabname}}--|>{\arrayrulecolor{black}}--}
				\rowcolor[rgb]{0.92,0.92,0.92} \multicolumn{1}{c}{\multirow{-2}{*}{\normalsize Method}}  & \multicolumn{1}{c||}{\multirow{-2}{*}{\normalsize Dataset}}   & {\normalsize 20\%} & {\normalsize 10\%} \\ 
				\hline \hline
				\multicolumn{4}{l}{\textit{\textbf{\normalsize Training-based  (0\% : 76.0)}}} \\[0.5ex]
				\hline 
				{\normalsize HydraViT}\pub{NeurIPS24} \cite{haberer2024hydravit}  & {\normalsize ADE20K} & {\normalsize 73.6} & {\normalsize 74.8}\\
				{\normalsize EA-ViT}\pub{ICCV25} \cite{zhu2025ea}  & {\normalsize ADE20K} & {\normalsize 74.9} &  {\normalsize 75.7} \\ \hline \hline
				\multicolumn{4}{l}{\textit{\textbf{\normalsize Training-free  (0\% : 76.0)}}} \\[0.5ex]
				\hline
				{\normalsize LAMP}\pub{ICLR21} \cite{lee2021layer}  & {\normalsize ImageNet}  &  {\normalsize 70.8} & {\normalsize 74.2}  \\
				{\normalsize NVIT}\pubbig{CVPR23}  \cite{yang2023global} & {\normalsize ImageNet}  & {\normalsize 70.7 } & {\normalsize 73.9 }  \\
				{\normalsize SNIP}\pub{ICCV23} \cite{kohama2023single}  & {\normalsize ImageNet}  & {\normalsize 71.7} &  {\normalsize 74.5}  \\
				{\normalsize SNOWS}\pub{ICLR25} \cite{LucasM25}   & {\normalsize ImageNet} & {\normalsize 72.9}  &  {\normalsize 74.1} \\
				{\normalsize SnapViT}\pub{NeurIPS25} \cite{simoncini2025elastic}  & {\normalsize ImageNet}  & {\normalsize 72.6} & {\normalsize 73.6}  \\ 
				\cdashline{1-4}
				\noalign{\vskip 1pt} 
				\rowcolor[rgb]{0.92,0.92,0.92}{\normalsize \textbf{\textsc{Cut-ViT} (Ours)}}   & {\normalsize ADE20K}  & \textbf{\normalsize 74.7}{\scriptsize \textbf{\textcolor{impro}{$\uparrow$1.8}}} & \textbf{\normalsize 75.2}{\scriptsize \textbf{\textcolor{impro}{$\uparrow$1.1}}}   \\ 
				\hline
				
			\end{tabular}
			\label{tab:ood}
		}
	\end{minipage}
\end{table*}

\begin{table*}[t]
	\centering
	\setlength{\tabcolsep}{2.7pt}
	\small
	\begin{minipage}[t]{0.55\textwidth}
		\centering
		\renewcommand\arraystretch{1.15}
		\setlength{\tabcolsep}{1.7pt}
		\caption{Quantitative results ($\S\ref{compare}$) for image classification on ImageNet \cite{deng2009imagenet}.}
		\vspace{-0.2cm}
		\resizebox{\textwidth}{!}{
			\begin{tabular}{r>{\centering\arraybackslash}p{1.8cm}||>{\centering\arraybackslash}p{1.4cm}|
					>{\centering\arraybackslash}p{1.4cm}|
					>{\centering\arraybackslash}p{1.4cm}}
				\thickhline
				\rowcolor[rgb]{0.92,0.92,0.92} &  &      \multicolumn{3}{c}{\normalsize Different Sparsities} \\  \hhline{>{\arrayrulecolor{tabname}}--|>{\arrayrulecolor{black}}---}
				\rowcolor[rgb]{0.92,0.92,0.92} \multicolumn{1}{c}{\multirow{-2}{*}{\normalsize Method}} & \multicolumn{1}{c||}{\multirow{-2}{*}{\normalsize Dataset}}   & {\normalsize 10\%} & {\normalsize 20\%} & {\normalsize 30\%} \\ 
				\hline \hline
				\multicolumn{5}{l}{\textit{\textbf{\normalsize Training-based  (0\% : 89.30)}}} \\[0.5ex]
				\hline 
				{\normalsize HydraViT} \pub{NeurIPS24} \cite{haberer2024hydravit}  & {\normalsize ImageNet} & {\normalsize 87.8} & {\normalsize 78.6} & {\normalsize 55.9}\\
				{\normalsize EA-ViT} \pub{ICCV25} \cite{zhu2025ea}  & {\normalsize ImageNet} & {\normalsize 89.1} & {\normalsize 81.7} & {\normalsize 60.3} \\ \hline \hline
				\multicolumn{5}{l}{\textit{\textbf{\normalsize Training-free (0\% : 89.30)}}} \\[0.5ex]
				\hline
				{\normalsize NVIT} \pubbig{CVPR23}  \cite{yang2023global} & {\normalsize ImageNet}  & {\normalsize 71.3 } & {\normalsize 61.9 }  & {\normalsize 54.7 }  \\
				{\normalsize SNIP} \pub{ICCV23} \cite{kohama2023single}  & {\normalsize ImageNet}  & {\normalsize 72.1} & {\normalsize 63.4} &  {\normalsize 55.2} \\
				{\normalsize SNOWS} \pub{ICLR25} \cite{LucasM25}  & {\normalsize ImageNet} & {\normalsize 86.2}  & {\normalsize 76.7} &  {\normalsize 56.1}\\
				{\normalsize SnapViT} \pub{NeurIPS25} \cite{simoncini2025elastic}  & {\normalsize ImageNet}  & {\normalsize 86.8} & {\normalsize 78.2} & {\normalsize 56.4}  \\
				\cdashline{1-5}
				\noalign{\vskip 1pt} 
				\rowcolor[rgb]{0.92,0.92,0.92}{\normalsize \textbf{\textsc{Cut-ViT} (Ours)}}   &{\normalsize ImageNet}  & \textbf{\normalsize 88.6} {\scriptsize \textbf{\textcolor{impro}{$\uparrow$1.8}}} & \textbf{\normalsize 79.4} {\scriptsize \textbf{\textcolor{impro}{$\uparrow$1.2}}}  & \textbf{\normalsize 57.2} {\scriptsize \textbf{\textcolor{impro}{$\uparrow$0.8}}}  \\
				\hline
		\end{tabular}}
		\label{tab:image_classification}
		\vspace{-0.2cm}
	\end{minipage}
	\hfill
	\begin{minipage}[t]{0.43\textwidth}
		\renewcommand\arraystretch{1.27}
		\setlength{\tabcolsep}{1.7pt} 
		\caption{Complexity analysis ($\S\ref{ablation_study}$) on a single A100 GPU under the same settings.}
		\vspace{-0.2cm}
		\resizebox{\textwidth}{!}{
			\begin{tabular}{r||c|c|c}
				\thickhline
				\rowcolor[rgb]{0.92,0.92,0.92} \multicolumn{1}{c||}{\normalsize Method}  & {\normalsize \makecell{Pruning Time \\ (Second)}} & {\normalsize \makecell{~GPU~ \\ (GB)}} & {\normalsize mIoU} \\ 
				\hline \hline
				\multicolumn{4}{l}{\textit{\textbf{\normalsize Training-based }}} \\[0.5ex]
				\hline
				{\normalsize EA-ViT} \pub{ICCV25} \cite{zhu2025ea}  & {\normalsize 13920} & {\normalsize 37.9} & {\normalsize 50.9}\\
				\hline \hline
				\multicolumn{4}{l}{\textit{\textbf{\normalsize Training-free }}} \\[0.5ex]
				\hline
				{\normalsize SNIP} \pub{ICCV23} \cite{kohama2023single}    & {\normalsize 220} & {\normalsize 27.0} & {\normalsize 47.6}  \\
				{\normalsize SNOWS} \pub{ICLR25} \cite{LucasM25}    & {\normalsize 6121}  & {\normalsize 21.3} & {\normalsize 47.8} \\
				{\normalsize SnapViT} \pub{NeurIPS25} \cite{simoncini2025elastic}   & {\normalsize 292} & {\normalsize 23.5} &  {\normalsize 48.3} \\
				\cdashline{1-4}
				\noalign{\vskip 1pt} 
				\rowcolor[rgb]{0.92,0.92,0.92}{\normalsize \textbf{\textsc{Cut-ViT} (Ours)}}     & {\normalsize \textbf{61}}& {\normalsize \textbf{10.7}}  & {\normalsize \textbf{50.0}} \\
				\hline
		\end{tabular}}
		\label{tab:complexity}
		\vspace{-0.2cm}
	\end{minipage}
	
\end{table*}

\noindent\textbf{Image Classification.}  Table \ref{tab:image_classification} presents the image classification results of \textsc{Cut-ViT} on the ImageNet dataset \cite{deng2009imagenet}, compared with state-of-the-art methods. Even without gram anchoring, the method outperforms training-free pruning approaches by relying solely on basis-agnostic and residual constraints. At sparsity levels of 10\%, 20\%, and 30\%, it achieves gains of \textbf{1.8\%}, \textbf{1.2\%}, and \textbf{0.8\%}, respectively. Furthermore, consistent with dense prediction tasks, \textsc{Cut-ViT} achieves competitive performance relative to training-based pruning methods. These results indicate that basis-invariant subspace consistency, even when applied solely to  CLS features, effectively captures rich global information.

\subsection{Out of Distribution Generalization}
\label{ood}
The generalization of \textsc{Cut-ViT} is evaluated by performing inference on the PASCAL VOC 2012 dataset \cite{everingham2015pascal} using models trained on ADE20K \cite{zhou2017scene}. Table \ref{tab:ood} compares its performance with state-of-the-art methods. \textsc{Cut-ViT} consistently outperforms training-free pruning approaches across all sparsity levels, achieving a notable gain of \textbf{1.5\%} mIoU at 20\% sparsity. These results provide strong evidence that the proposed pruning strategy enables the subnetworks to effectively inherit the generalization ability of the native DINOv3 \cite{simeoni2025dinov3}.

\subsection{Ablation Study}
\label{ablation_study}

\noindent\textbf{Complexity Analysis.}  Table \ref{tab:complexity} presents a comparison of the algorithmic complexity of \textsc{Cut-ViT} with state-of-the-art methods. Compared with both training-free and training-based methods, \textsc{Cut-ViT} achieves promising performance while significantly reducing pruning time and GPU memory usage. Specifically, it requires only \textbf{20.9\%} of the  time and \textbf{45.5\%} of the GPU memory required by SnapViT \cite{simoncini2025elastic}, while delivering a \textbf{1.7\%} improvement in mIoU. Although the method trails training-based pruning approaches, such as EA-ViT \cite{zhu2025ea}, by 0.9\% in performance, it requires only \textbf{0.44\%} of the time and \textbf{28.2\%} of the GPU memory. These results demonstrate that competitive performance can be achieved with minimal computational overhead.

\noindent\textbf{Component Analysis.} The results for each component and the different gram matrices are reported in Tables \ref{tab:sub_space_residual_De} and \ref{tab:gram}, respectively. Local dense feature alignment $\Lcal_{ba}$ is adopted as the baseline.  The basis-agnostic constraint, residual constraint, and task-specific adaptation mutually complement each other, enabling the model to achieve excellent experimental performance. Furthermore, incorporating spatial and channel gram matrices encourages the pruned model to learn spatial contextual information and channel-wise semantic representations aligned with those of DINOv3 \cite{simeoni2025dinov3}. The complementary effects of the two gram matrices allow the model to accurately capture semantic and spatial contextual information, thereby achieving optimal performance.

\begin{table*}[t]
	\centering
	\renewcommand\arraystretch{1.2}
	\setlength{\tabcolsep}{2.7pt}
	
	\begin{minipage}[t]{0.52\textwidth}
		\setlength{\tabcolsep}{2.5pt}
		\centering
		\scriptsize
		\caption{Component analysis ($\S\ref{ablation_study}$) of basis-agnostic, residual constraints and task-specific on ADE20K \cite{zhou2017scene} with 20\% sparsity.}
		\vspace{-0.2cm}
		\resizebox{\textwidth}{!}{
			
			\begin{tabular}{cccc||c}
				\thickhline
				\rowcolor[rgb]{0.92,0.92,0.92} Baseline & \makecell{Basis-agnostic \\ Constraint}    & \makecell{Residual \\ Constraint} & \makecell{Task- \\specific}  & mIoU   \\
				\hline \hline
				$\checkmark$ & & &  & 48.4\\
				$\checkmark$ & $\checkmark$ & & & 49.1 \\
				$\checkmark$ &  & $\checkmark$ & & 49.3 \\
				\cdashline{1-4}
				\noalign{\vskip 1pt} 
				\rowcolor[rgb]{0.92,0.92,0.92} $\checkmark$ & $\checkmark$ & $\checkmark$ &  $\checkmark$ & \textbf{49.7} \\
				\hline
			\end{tabular}
			\label{tab:sub_space_residual_De}
		}
	\end{minipage}
	\hfill
	\begin{minipage}[t]{0.45\textwidth}
		\centering
		\renewcommand\arraystretch{1.27}
		\setlength{\tabcolsep}{2.5pt}
		\scriptsize
		\caption{Component analysis ($\S\ref{ablation_study}$) of spatial and channel gram anchoring on ADE20K \cite{zhou2017scene} with 20\% sparsity.}
		\vspace{-0.2cm}
		\resizebox{\textwidth}{!}{
			\begin{tabular}{ccc||c}
				\thickhline
				\rowcolor[rgb]{0.92,0.92,0.92} Baseline & \makecell{Spatial Gram \\  Anchoring} & \makecell{Channel  Gram \\  Anchoring} & mIoU   \\
				\hline \hline
				$\checkmark$ & & &  48.4\\
				$\checkmark$ & $\checkmark$ & & 49.7 \\
				$\checkmark$ &  & $\checkmark$ & 49.5 \\
				\cdashline{1-4}
				\noalign{\vskip 1pt} 
				\rowcolor[rgb]{0.92,0.92,0.92} $\checkmark$ & $\checkmark$ & $\checkmark$ & \textbf{50.0} \\
				\hline
			\end{tabular}
			\label{tab:gram}
		}
	\end{minipage}
\end{table*}

\begin{table*}[t]
	\centering
	\renewcommand\arraystretch{1.15}
	\setlength{\tabcolsep}{2.7pt}
	
	\begin{minipage}[t]{0.52\textwidth}
		\centering
		\renewcommand\arraystretch{1.15}
		\setlength{\tabcolsep}{3.5pt}
		\scriptsize
		\caption{Parameter analysis ($\S\ref{ablation_study}$) of principal component number $K$ on ADE20K \cite{zhou2017scene} at 20\% sparsity. CEVR denotes the cumulative explained variance ratio.}
		\vspace{-0.2cm}
		\resizebox{\textwidth}{!}{
			\begin{tabular}{c||ccccccc}
				\stickhline
				\rowcolor[rgb]{0.92,0.92,0.92} $K$ & 96  & 128  & 144  & 192 & 256 & 300 & 500    \\
				\hline \hline
				CEVR & 91.8 & 93.9 & 96.6 & 98.9 & 99.2 & 99.4 & 99.6 \\
				Time ($10^{-1}s$)& 0.43 & 0.46 & 0.5 & 0.59 & 0.72 & 0.80 & 1.12 \\
				\hline 
				Classification & 79.1 & 79.2 & 79.2 & 79.4 & 79.4 & 79.4 & 79.5 \\ 
				Segmentation & 49.6 & 49.8 & 49.9 & 50.0 & 50.1 & 50.1 & 50.2\\
				\hline
			\end{tabular}
			\label{tab_k_component}
			\vspace{-0.5cm}
		}
	\end{minipage}
	\hfill
	\begin{minipage}[t]{0.45\textwidth}
		\setlength{\tabcolsep}{0.3pt}
		\renewcommand\arraystretch{1.2}
		\centering
		\scriptsize
		\caption{Performance analysis ($\S\ref{ablation_study}$) between static weighting and spectral entropy weighting across different tasks at 30\% sparsity.}
		\vspace{-0.2cm}
		\resizebox{\textwidth}{!}{
			
			\begin{tabular}{c||c|c}
				\stickhline
				\rowcolor[rgb]{0.92,0.92,0.92}
				\raisebox{1.2ex}{Setup} & \shortstack{Static\\Weighting} 
				& \shortstack{\rule{0pt}{2.1ex} Spectral Entropy\\Weighting}   \\
				\hline \hline
				 Detection & 47.9 & \textbf{48.6}\\
				   Segmentation & 58.8 / 57.2 / 60.3 & \textbf{59.4} / \textbf{58.0} / \textbf{60.8}\\
				  Matching & 70.2 / 55.5 / 37.3 & \textbf{70.8} / \textbf{56.3} / \textbf{37.9}\\
				\hline
			\end{tabular}
			\label{tab:weighting}
			\vspace{-0.5cm}
		}
	\end{minipage}
\end{table*}

\noindent\textbf{Parameter \& Weighting Analysis.}  Tables \ref{tab_k_component} and \ref{tab:weighting} present the performance analysis of the number of principal components $K$ and different loss weightings, respectively. For the former, we analyze it from the following three aspects: \textbf{i)} Information preservation.
We introduce the \textit{cumulative explained variance ratio} (CEVR) to quantify the information retained by the selected Top-\textit{K} principal components. \textbf{ii)} Efficiency. We measure the time to obtain the corresponding principal components from a single matrix via SVD. \textbf{iii)} Cross-task robustness. The results below confirm that $K = 192$ achieves an optimal trade-off among information preservation, computational efficiency, and performance. For the latter, the performance on object detection, video object segmentation, and semantic matching highlights the superiority of spectral entropy weighting.

\noindent \textbf{Pruning Dataset Analysis.} We conduct an ablation analysis on pruning with different datasets across various tasks, as shown in Table \ref{tab:pruning_dataset_analysis}. The experimental results on the three tasks show that even when pruning is performed on the corresponding target datasets, the state-of-the-art SNOWS \cite{LucasM25} and SnapViT \cite{simoncini2025elastic} achieve only marginal performance gains. This directly demonstrates that the performance improvements stem from the superiority of our method itself, rather than from simply replacing the pruning dataset.

\noindent \textbf{Architecture Robustness Analysis.} In addition to DINOv3 \cite{simeoni2025dinov3}, we further apply our method to SAM \cite{kirillov2023segment}, DeiT \cite{touvron2021training}, and CLIP \cite{radford2021learning}, and compare it against state-of-the-art methods to validate its robustness across architectures, as reported in Table \ref{tab:robutness_network}. Their performance on prompt segmentation, video object segmentation, and open-vocabulary segmentation is evaluated using 0.1\% of SA-1B, DAVIS-2017, and PASCAL-Context, respectively. The results demonstrate that our method generalizes robustly across diverse architectures.

\begin{figure}[t]
	\centering
	\includegraphics[width=1\linewidth]{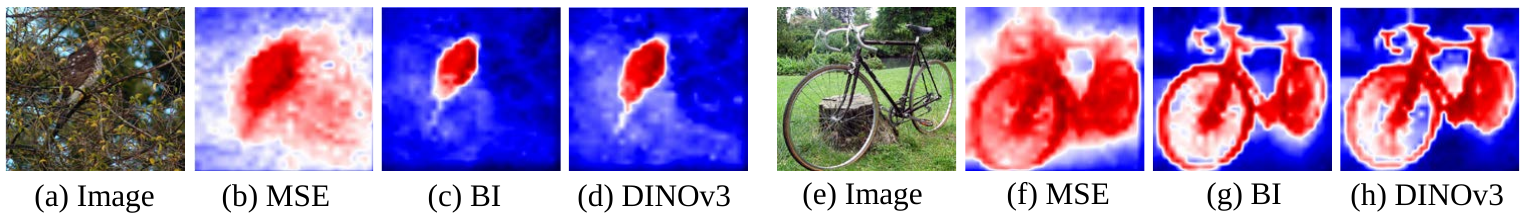}
	\vspace{-0.5cm}
	\caption{Qualitative comparison ($\S\ref{ablation_study}$) of different subspace alignment strategies.}
	\label{heat_map}
	\vspace{-0.1cm}
\end{figure}

\noindent \textbf{Subspace Alignment Analysis.} Table \ref{tab:mse_bi} and Fig. \ref{heat_map} provide an extensive analysis of subspace alignment strategies from quantitative and qualitative perspectives, respectively. Taken together, these results show that naively aligning and pruning models with standard distance metrics (\textit{e.g.}, MSE) penalizes meaningless rotational differences, leading to ill-posed optimization. In contrast, our basis invariance (BI) proves that $\mathcal{L}_{basis}$ depends exclusively on the geometric overlap between subspaces and is invariant to basis rotations, thereby providing a robust foundation for subspace alignment. Compared with the diffused representations induced by MSE, BI preserves the precise object boundaries of DINOv3 \cite{simeoni2025dinov3} representations and achieves superior performance.

\begin{table*}[t]
	\begin{center}
		\renewcommand\arraystretch{1.2}
		\caption{Ablation analysis ($\S\ref{ablation_study}$) on different pruning datasets at 30\% sparsity.}
		\vspace{-0.2cm}
		\setlength{\tabcolsep}{0.6pt}
		\resizebox{1.0\textwidth}{!}{
			{\scriptsize
				\begin{tabular}{>{\raggedleft}p{3.4cm}>{\centering\arraybackslash}p{1.6cm}||c|c|c}
					\thickhline
				\rowcolor[rgb]{0.92,0.92,0.92} &  &  Video Object Segmentation  & Depth Estimation  & Semantic Matching \\ \hhline{>{\arrayrulecolor{tabname}}--|>{\arrayrulecolor{black}}---}
				\rowcolor[rgb]{0.92,0.92,0.92} \multicolumn{1}{c}{\multirow{-2}{*}{Method}}  & \multicolumn{1}{c||}{\multirow{-2}{*}{Dataset}}   & $(\Jcal\&\Fcal)_m$ / $\Jcal_m$ / $\Fcal_m$ 	& ARel \textcolor{red}{\textbf{$\downarrow$}} / $\delta_1$ \textcolor{red}{\textbf{$\uparrow$}} 
				& FG3DCar / JODS / SBD \\ 
				\hline \hline
				  &  ImageNet  &  52.9 / 50.7 / 55.1 & 8.3 / 97.6 &  64.8 / 44.2 / 32.1 \\
				     \multirow{-2}{*}{SNOWS \pub{ICLR25} \cite{LucasM25}} & Target &  53.4 / 50.9 / 56.0 & 8.0 / 97.6  & 65.4 / 44.6 / 32.6 \\
				     \hline 
				  &  ImageNet  &  56.0 / 54.3 / 57.7  & 7.7 / 97.6 &  65.7 / 43.3 /32.4  \\
				 \multirow{-2}{*}{SnapViT \pub{NeurIPS25} \cite{simoncini2025elastic}} & Target & 56.4 / 54.6 / 58.1  &  7.3 / 97.7   & 66.4 / 43.5 / 32.8 \\
				 \hline 
					
				\textbf{\textsc{Cut-ViT} (Ours)}~~~    & Target & \makecell{\textbf{59.4} / \textbf{58.0} / \textbf{60.8} \\ {\tiny \textbf{\textcolor{impro}{$\uparrow$3.0}}}~/ {\tiny \textbf{\textcolor{impro}{$\uparrow$3.4}}} / {\tiny \textbf{\textcolor{impro}{$\uparrow$2.7}}}}  & \makecell{ \textbf{4.6} / \textbf{98.7} \\ {\tiny \textbf{\textcolor{impro}{$\uparrow$2.7}}}~/ {\tiny \textbf{\textcolor{impro}{$\uparrow$1.0}}}}  & \makecell{\textbf{70.8}  / \textbf{56.3} / \textbf{37.9} \\ {\tiny \textbf{\textcolor{impro}{$\uparrow$4.4}}}~/ {\tiny \textbf{\textcolor{impro}{$\uparrow$11.7}}} / {\tiny \textbf{\textcolor{impro}{$\uparrow$5.1}}}} \\
					
					\hline
				\end{tabular}
			}
		}
		\label{tab:pruning_dataset_analysis}
		\vspace{-0.4cm}
	\end{center}
\end{table*}

\begin{table*}[t]
	\centering
	\setlength{\tabcolsep}{1.7pt}
	\small
	\begin{minipage}[t]{0.64\textwidth}
		\centering
		\renewcommand\arraystretch{1.4}
		\setlength{\tabcolsep}{0.3pt}
		\caption{Robustness analysis ($\S\ref{ablation_study}$) of different architectures on their corresponding tasks at 30\% sparsity.}
		\vspace{-0.2cm}
		\resizebox{\textwidth}{!}{
			\begin{tabular}{c||c|c|c}
				\thickhline
				\rowcolor[rgb]{0.92,0.92,0.92} & {\large  SAM }  & {\large DeiT }  & {\large CLIP } \\ \hhline{>{\arrayrulecolor{tabname}}-|>{\arrayrulecolor{black}}---}
				\rowcolor[rgb]{0.92,0.92,0.92} \multicolumn{1}{c||}{\multirow{-2}{*}{\large Method}}  &   {\large Promptable Seg.} 	& {\large Video Object Seg.}
				& {\large Open-vocabulary Seg.} \\
				\hline \hline
			{\large	SNOWS \cite{LucasM25}}  & {\large 67.9}  & {\large 40.1} / {\large 42.3} /{\large 37.8} & {\large 45.7} \\
				{\large SnapViT  \cite{simoncini2025elastic}} & {\large 68.6} & {\large 40.6} / {\large 43.0} / {\large 38.1} & {\large 46.9} \\ \hline
				{\large \textbf{\textsc{Cut-ViT}}} & {\large \textbf{71.3}} {\footnotesize \textbf{\textcolor{impro}{$\uparrow$2.7}}} & {\large \textbf{44.6}} {\footnotesize \textbf{\textcolor{impro}{$\uparrow$4.0}}} / {\large \textbf{46.7}} {\footnotesize \textbf{\textcolor{impro}{$\uparrow$3.7}}} / {\large \textbf{42.5}} {\footnotesize \textbf{\textcolor{impro}{$\uparrow$4.4}}} & {\large \textbf{51.1}} {\footnotesize  \textbf{\textcolor{impro}{$\uparrow$3.2}}} \\
			   \hline
				
		\end{tabular}}
		\label{tab:robutness_network}
		\vspace{-0.2cm}
	\end{minipage}
	\hfill
	\begin{minipage}[t]{0.33\textwidth}
		\renewcommand\arraystretch{0.95}
		\setlength{\tabcolsep}{1.1pt} 
		\caption{Ablation analysis ($\S\ref{ablation_study}$) between MSE and BI.}
		\vspace{-0.2cm}
		\resizebox{\textwidth}{!}{
			\begin{tabular}{>{\centering\arraybackslash}p{1.3cm}
					>{\centering\arraybackslash}p{0.8cm}
					>{\centering\arraybackslash}p{0.8cm}|
					>{\centering\arraybackslash}p{0.8cm}}
				\thickhline
				\rowcolor[rgb]{0.92,0.92,0.92}  {\scriptsize Baseline} &  {\scriptsize MSE} &  {\scriptsize BI} &  {\scriptsize mAP} \\
				  \hline
				 	$\checkmark$ & & & {\scriptsize 54.3}  \\
				 		$\checkmark$ & $\checkmark$ & & {\scriptsize 54.7}  \\
				 		$\checkmark$ &  & $\checkmark$ & {\scriptsize \textbf{55.7}}  \\
				\hline
		\end{tabular}}
		\label{tab:mse_bi}
		\vspace{-0.2cm}
	\end{minipage}
	
\end{table*}

\begin{figure}[t]
	\centering
	\includegraphics[width=1\linewidth]{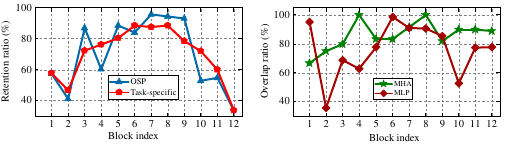}
	\vspace{-0.5cm}
	\caption{The parameter retention and overlap analysis ($\S\ref{ablation_study}$) for each block in DINOv3 \cite{simeoni2025dinov3}. (\textbf{Left}) shows the parameter retention rate for each block in both training-free OSP and our task-specific methods. (\textbf{Right}) illustrates the overlap ratio of the retained parameters in the MLP and MHA layers of each block for the two types of methods.}
	\label{fig3_param_compare}
	\vspace{-0.1cm}
\end{figure}

\noindent\textbf{Retention $\&$ Overlap Analysis.}  Fig. \ref{fig3_param_compare} shows the parameter retention and overlap ratios for both training-free OSP and our task-specific methods across different blocks. As illustrated in (\textbf{Left}), the parameter retention rate of the generic method fluctuates significantly from the 3rd to the 10th layer, while our task-specific method exhibits more stable changes, indicating that our approach better preserves the fine-grained texture features of DINOv3 \cite{simeoni2025dinov3}. The curve in (\textbf{Right}) reveals that both pruning paradigms focus more on the information within the attention layers across several intermediate blocks.

\section{Conclusion}
In this paper, we propose \textsc{Cut-ViT}, a task-specific model pruning paradigm to mitigate robustness degradation and task-specificity deficiency present in existing work. Specifically, we construct gram anchoring matrices from spatial and channel perspectives and apply subspace decomposition to extract the corresponding bases in latent space. Basis-agnostic and residual constraints are designed to enforce subspace consistency between the original and pruning-oriented DINOv3 models across both domains. This enables the subnetwork to inherit the robust feature representations of foundation models such as DINOv3. Moreover, we introduce spectral entropy adaptation, which quantifies the information density of feature manifolds along spatial and channel dimensions, thereby adapting the pruning objective to specific downstream tasks.
Extensive experiments on \textbf{six} tasks across \textbf{nine} datasets show that \textsc{Cut-ViT} achieves state-of-the-art performance with minimal computational overhead compared to prior approaches.

\begin{sloppypar}
\noindent \textbf{Acknowledgments.} This work was supported by the National Defense Science and Technology Industry Bureau Technology Infrastructure Project (JSZL2024606C001).
\end{sloppypar}
%
%
\bibliographystyle{splncs04}
\bibliography{main}
\end{document}